# Font Identification in Historical Documents Using Active Learning

Anshul Gupta[1], Ricardo Gutierrez-Osuna[1], Matthew Christy[2], Richard Furuta[1], Laura Mandell[2]

[1]Department of Computer Science and Engineering, Texas A&M University
[2]Initiative for Digital Humanities, Media, and Culture, Texas A&M University
{anshulg, rgutier, mchristy, furuta, mandell}@tamu.edu

**Abstract**

Identifying the type of font (e.g., Roman, Blackletter) used in historical documents can help optical character recognition (OCR) systems produce more accurate text transcriptions. Towards this end, we present an active-learning strategy that can significantly reduce the number of labeled samples needed to train a font classifier. Our approach extracts image-based features that exploit geometric differences between fonts at the word level, and combines them into a bag-of-word representation for each page in a document. We evaluate six sampling strategies based on uncertainty, dissimilarity and diversity criteria, and test them on a database containing over 3,000 historical documents with Blackletter, Roman and Mixed fonts. Our results show that a combination of uncertainty and diversity achieves the highest predictive accuracy (89% of test cases correctly classified) while requiring only a small fraction of the data (17%) to be labeled. We discuss the implications of this result for mass digitization projects of historical documents.

## Introduction

Digitization provides easy access to most of the documents published in the modern era, from texts to images and video. By comparison, printed historical documents—everything from pamphlets to ballads to multi-volume poetry collections in the hand-press period (roughly 1475-1800)—are difficult to access by all but the most devoted scholars. The need to create machine-searchable collections has accelerated work on Optical Character Recognition (OCR) of historical documents.

OCR of historical documents is a challenging task, partly due to the physical integrity of the documents and the quality of the scanned images, but also because of the font characteristics. Historical documents in the hand-press period have irregular fonts, and show large variations within a single font class since the early printing processes had not been standardized. Blackletter (or Gothic) and Roman font classes are the two main font types used in early modern printing, but these two font classes have evolved into multiple subclasses since the first printed book. Knowing the font type and characteristics for each document in a collection can substantially improve the performance of OCR systems (La Manna et al. 1999, Imani et al. 2011). In large collections, however, hand-tagging each individual document, page and text region according to its font becomes prohibitive. As an example, the Eighteenth Century Collections Online (ECCO) and Early English Books Online (EEBO) databases –two of the largest collections available—contain over 45 million pages.

To address this problem, we present a font-identification system that can be used to automatically tag individual documents within a large collection according to their fonts[1]. Font identification is best formulated as a supervised classification problem, and as such it requires labeled data for model building. Classification models work best when they have sufficient labelled data that represent the diversity of exemplars in the corpus. In our case, however, obtaining large amounts of labeled data from a corpus of 45 million page images, with varied font types, is a daunting task. For this reason, we propose an active-learning approach to optimize the hand labeling process. Active learning is a mixed-initiative paradigm where a machine learning (ML) algorithm and a human work together during model building: the ML algorithm suggests a few high-value unlabeled exemplars, these are passed to the human to obtain labels, the model is adapted based on these newly-labeled exemplars, and the process is repeated until the model converges.

The remaining parts of this document are organized as follows. First, we review the characteristics of historical fonts and how they may be exploited for automated classification. Next, we describe the proposed active-learning

---

methodology, including the feature extraction process, the sampling strategies used to select informative unlabeled instances, and the classification model. Then, we present an experimental comparison on a database containing over 3,000 documents with Roman and Blackletter fonts. Our results show that active learning can achieve a classification accuracy of 89% on test data using a small fraction (17%) of the training corpus. The article concludes with a discussion of these results and directions for future work.

## Background and Related work

### Historical fonts and their characteristics

Historical fonts can be broadly categorized in one of two categories: Blackletter or Roman. The name Blackletter refers to a highly ornamental script style of calligraphy that was widely used in Europe between the $12^{th}$ and $17^{th}$ centuries. To make printed books look similar to hand-lettered, Johann Gutenberg created font sets based on Blackletter (or Gothic) style for the first printing press between 1439 and 1455. Blackletter fonts have tall, narrow letters, with sharp, straight, angular lines. Characters typically do not connect with each other, especially in round letters (see Figure 1a-b). Finally, line strokes also drastically differ in thickness, and the between-letter spacing is very small, which makes text printed with this font quite difficult to read.

Compared to Blackletter, Roman typefaces are significantly less ornamental. The first Roman typefaces were designed around 1460 by Nicholas Jenson with the intention of making an easier-to-read font. Roman fonts take space proportional to their shape, put less emphasis on angled strokes, and possess serifs. They have curved strokes, which are also the thinnest parts of a character. In summary, angled strokes, between-letter spacing, and stroke width distribution provide information to discriminate between these two broad types of fonts.

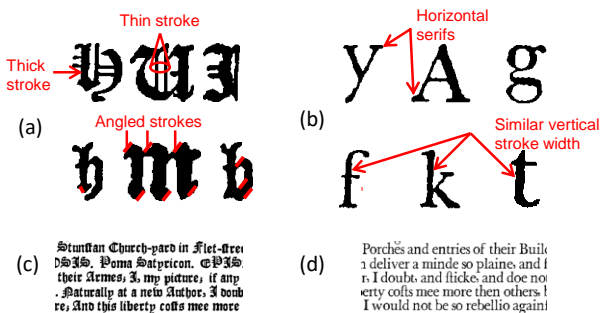

Figure 1: (a-b) Characteristics of Blackletter and Roman font; (c-d) Corresponding text snippets.

### Related work

OCR systems are divided into three groups according to the number of fonts present in the document collection: mono-font, multi-font, and omni-font systems. As the name suggests, mono-font OCR systems deal with a single font; as a result, they tend to have high recognition accuracy. In contrast, omni-font OCR systems can recognize text printed in any font, but they generally show very poor recognition accuracy. Finally, multi-font OCR systems work on a set of fonts and can produce good results provided the font classes have good separability.

OCR systems work well with modern documents, but not for historical documents. The fonts and layout of the historical documents vary substantially from one document to other, which makes training a single high-performance OCR engine difficult. Ait-Mohand et al. (2010) proposed an HMM-based multi-font OCR system that modifies its HMM model for each document to better recognize a specific font class. When tested on a dataset with 17 modern font classes, the authors reported a 98% recognition rate, which is very close to the recognition rates of single-font OCR systems (99%). An alternative to adapting a multi-font OCR model is to train multiple mono-font OCR systems and direct documents to the respective OCR engine. This requires, however, that the font class of each document be identified in advance.

In an extensive literature survey, Ghosh et al. (2010) organize font recognition methodologies into two categories based on the extracted features: structure-based, and appearance-based. Structure-based methods extract the connected components of characters, and analyze their shape and structure to recognize particular fonts. In contrast, appearance-based methods use features that can capture the visual appearance of the individual characters and the way they are grouped into words, lines and paragraphs. These two categories are further divided according to whether they operate at the document, page, paragraph, or word level. Rani et al. (2013) presented a character-level font identifier. Using Gabor and gradient features and an SVM-classifier, they reported 99% average accuracy on a dataset with 19,000 images of characters and numerals from 17 fonts for the English language, and 14 fonts for the Gurmukhi language.

Whether they are used to classify documents, words, or characters, font identification systems incur a large cost to generate sufficient labeled data to train the classifier. Recent studies, however, have shown that active learning can be very effective in reducing the large overhead of labeling data (Schohn et al. 2000, Fu et al. 2013). Unsurprisingly, active learning has begun to garner attention in the document analysis community. For example, Bouguelia et al. (2013) proposed a semi-supervised active learning algorithm for stream-based classification of documents into multiple classes, such as bank checks, medical receipts, invoices, or prescriptions. Compared to a model built with a fully labeled training dataset, active learning provided a 2-3% precision boost while using on average only 36% of the labeled data.

# Methods

Figure 2 illustrates the overall font-identification system. The process starts by selecting a set of seed training images from the ECCO/EEBO collection, which are then passed through an OCR system –in our case, the open-source Tesseract engine (Smith 2007). OCR outputs generally contain a number of noisy bounding boxes (BB) around non-textual elements, such as pictures, decorative elements, or bleed through. These noisy BBs can be problematic for font-identification since they behave as outliers when computing features. For this reason, as a first step we apply a de-noising algorithm (Gupta et al. 2015) to eliminate noisy BBs and return only those likely to contain text.

Once text BBs have been identified, they undergo an image preprocessing step that normalizes them by size (to a height of 400 pixels), filters out salt-and-pepper noise (Gonzalez et al. 2007) and removes skew (Kavallieratou et al. 2002). Preprocessed word images are then passed to a feature extraction module, to be described in the next section–see Table 1. The resulting feature vectors (from all the word images in a page) are then vector quantized and combined into a bag-of-words feature (BoF) representation for the page. These BoFs become the input to the font classifier.

Starting with a small set of seed training images, we iteratively train the font classifier and then use it to select the most informative (yet unlabeled) documents for the human analyst to label next. This training-labeling process (active learning) is repeated until a performance criterion (e.g., a target precision/recall rate) is met.

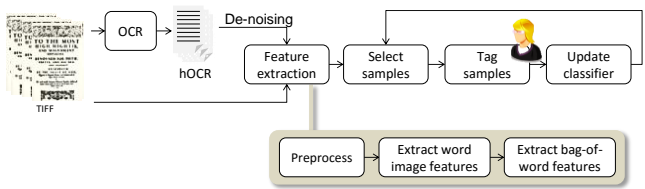

*Figure 2: Steps for active learning based font identification.*

## Word-level feature extraction

As summarized in Table 1, we extract three kinds of features to capture font information at the word-image level: the stroke width of each letter, the density of slanted lines, and the visual appearance of the word.

### 1. Average and IQR stroke width

We characterize the stroke width by its trimmed mean and IQR (interquartile range) on each word image. Namely, we scan 10% of the rows (41 rows; middle row ± 20 rows) of each preprocessed word image, and locate transitions from background to foreground, and transitions from foreground to background –see green and red points in Figure 3a, respectively. Difference in their x-coordinates serves as an estimate of the stroke width. Since each row may have multiple such stroke widths, we store the count $C_i$ for each row, and then compute the maximum $C_{max}$. Next, we select rows where $C_i = C_{max}$, and calculate the trimmed mean and IQR over their respective stroke widths. Using these robust statistics and limiting the computation to rows with $C_i = C_{max}$ provides further immunity to outliers.

### 2. Slant line density

We estimate the number of slanted lines by applying the Canny edge detector (Gonzalez et al. 2007) to each word image, followed by the Hough transform (Illingworth et al. 1988). We then compute the number of lines with slope in the range $45° \pm 5°$ and $-45° \pm 5°$, and divide it by the number of recognized characters in the word, which is available from the output of the OCR engine. This results in an estimate of the slant line density for a word image— see Figure 3b.

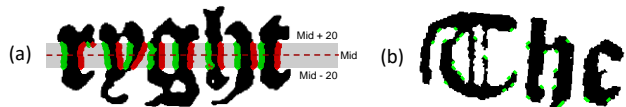

*Figure 3: (a) Calculating mean and IQR character width; (b) Results from the Hough transform; green line segments indicate the detected angled straight lines.*

### 3. Zernike moments

Finally, we capture the visual appearance of each word using Zernike Moments (ZMs) (Khotanzad et al. 1990). ZMs are the projection of the image onto an orthogonal basis known as the Zernike polynomials (ZPs):

$$Z_{m,n} = \frac{n+1}{\pi} \sum_x \sum_y I(x,y) V_{nm}^* \quad (1)$$

where $I(x,y)$ is the binary image, $V_{nm}^*$ are the ZPs, and $m$ and $n$ are the order of the ZMs. Following Tahmasbi et al. (2011), we compute the magnitude of first 6 ZMs along with their transformations (a total of 15 features).

*Table 1: Features extracted from each word image*

| Features | Justification |
|---|---|
| Average (trimmed) stroke width | Roman fonts have smaller vertical stroke width than blackletter fonts |
| IQR stroke width | Captures drastic differences in stroke width, typical of black letters |
| Slant line density | Blackletters fonts are characterized by angled lines and serifs |
| Zernike moments | Captures the overall shape (visual appearance) of each font |

## Page-level feature extraction: Bag-of-words model

Bag-of-word features (BoFs) (Sivic et al. 2003) are widely used to capture semantic information in images. BoFs are usually extracted using small patches in the image, which in our case correspond to the word images. To obtain a BoF for each page, we apply *k*-means (*k*=20) to the distribution of local features (across all training documents), vector quantize each word image, and compute the number of words assigned to each cluster. To achieve word-count invariance, we normalize each BoFs by the total number of words on the page.

**Active learning for font identification**

Active learning comprises of two parts: a base classifier and a sampling engine. The base classifier is trained on a small amount of labeled data ($L$), and the sampling engine uses it to select a batch of the most informative instances ($X$) from an unlabeled set ($U$) for labelling. A human annotator labels all instances in $X$, and these are added to $L$ to retrain the classifier. The entire process of training and sampling is repeated until convergence.

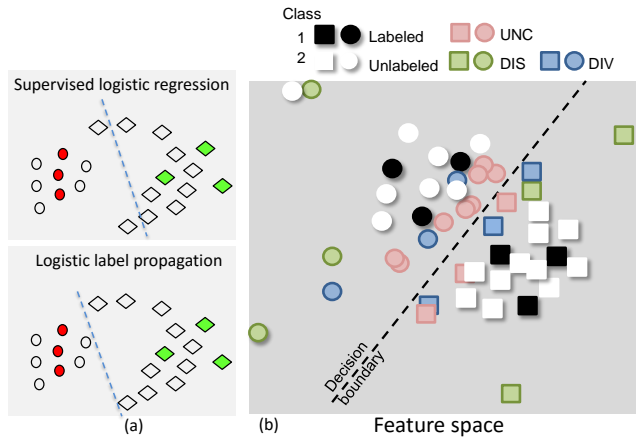

*Figure 4: (a) Example decision boundaries produced by logistic regression and LLP (Kobayashi et al. 2012). Filled markers represent labeled instances; (b) Illustration of the different active sampling criteria.*

*1. Base classifier*

We use a modified label-propagation model proposed by Kobayashi et al. (2012) known as Logistic Label Propagation (LLP). In label propagation (LP), all training instances (labeled+unlabeled) are treated as nodes in a fully connected graph, and labels are propagated to unlabeled data points according to their proximity to the labeled data:

$$S_{ij} = \exp(-\frac{||X_i - X_j||^2}{\sigma^2}) \quad (2)$$

where $X_i$ and $X_j$ are the features vectors for document $i$ and $j$, respectively, and $\sigma$ is the bandwidth hyper-parameter. A major drawback of LP is its computational complexity during recall; LP must reconstruct the whole similarity matrix for new instances, and then re-estimate their class-posteriors to predict their labels. In contrast, LLP trains a logistic classifier in a semi-supervised fashion by adding to the logistic cost function a cost derived from the LP model, which shifts the decision boundary to account for the distribution of unlabeled data –see Figure 4a.

*2. Active sampling*

The performance of active learning depends heavily on the choice of sampling strategy (known as a query function) that selects the most informative instances for labeling. For this reason, our query function considers three separate criteria:

a) **Uncertainty**. Following Settles (2012), unlabeled instances are selected according to the classifier's uncertainty about their labels –see Figure 4b. In particular, we compute uncertainty for an unlabeled data point $U_k$ as the entropy of its class-posterior distribution $p(y|U_k,L)$:

$$H(y|U_k,L) = -\sum_y p(y|U_k,L) \log p(y|U_k,L) \quad (3)$$

where $L$ is the labeled data and $y$ is the label variable that ranges over all possible labeling of $U_k$.

b) **Dissimilarity to the labeled data**. To promote exploration, we also consider instances that lie in unexplored regions of feature space; see Figure 4b. For each unlabeled instance ($U_k$), we find the 5 most-similar labeled instances ($L_n$) based on LLP's similarity matrix ($S_{nk}$). Samples with highest dissimilarity $D_k$ are selected for querying:

$$D_k = \frac{1}{5}\sum_{n=1}^{5} 1 - S_{nk} \quad (4)$$

c) **Diversity.** At each iteration, our sampling engine selects a batch of 20 unlabeled instances. To prevent instances within a batch from being too similar to each other, we incorporate a diversity metric that constructs the batch $X$ as follows:
 (1) Initialize $X$ with the unlabeled instance that has the largest score: $H(y|U_k) + D_k$
 (2) For each remaining unlabeled instance ($U_k$), calculate its diversity factor ($D'_k$) as:

$$D'_k = \min_n (1 - S_{nk}) \quad (5)$$

 where $S_{nk}$ is the similarity between $U_k$ and the $n^{th}$ sample already in $X$–see eq. (2)
 (3) Select sample $U_k$ with the largest combined score ($H(y|U_k) + D_k + D'_k$), and add it to $X$.
 (4) Repeat steps (2) and (3) above until 20 unlabeled instances have been selected.

**Results**

To evaluate the effectiveness of our font classifier, we devised two experiments to evaluate our feature set (Table 1) and determine the best active learning query function. In a first experiment, we examined whether the proposed fea-tures can discriminate between Blackletter and Roman fonts at the word level. In a second experiment, we evaluated the complete system using six possible active sampling strategies derived from combinations of three query criteria described in the previous section. For these experiments, we created a dataset consisting of 3,272 document images from the ECCO/EEBO databases: 1,005 printed in Blackletter, 1,768 in Roman, and 498 with text in both fonts (Mixed). Each of these documents was hand labeled by domain experts on the eMOP team.

**Evaluating word-image features**

To compare the discriminatory power of the different word-level feature sets, we randomly selected 500 Blackletter documents and 500 Roman documents, and extracted the 18 features shown in Table 1. We then trained a logistic regression classifier to predict the class label for each word. The threshold for the classifier output was selected by maximizing the accuracy through 5-fold cross-validation. Results are summarized in Figure 5a. The slant angle density and character width (SLD-CW) feature set achieves 58% classification rate, whereas the *ZM* feature set achieves 81%. When the two feature sets are combined (*ALL*), classification performance improves modestly to 84%. These results indicate that all the extracted features are important for font identification, and that they provide high between-class separability –even at the word level.

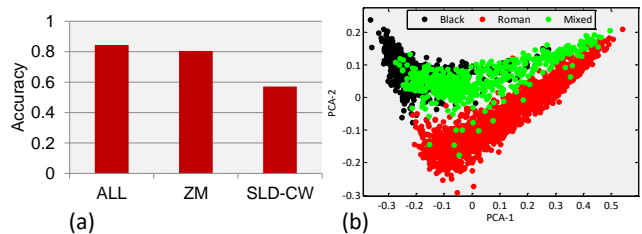

*Figure 5: (a) Cross-validated accuracy for classifiers trained using 18 features (ALL), Zernike moments (ZM), and slant line density and character width (SLD-CW); (b) Principal components analysis of the dataset using page-level features (BoFs); each point represents a document.*

**Evaluating the active learning system**

To evaluate the overall system, we randomly selected 600 documents (200 per class) as a test set, and used the remaining 2,672 documents as the training set. From these, we randomly selected 3 labeled documents (one per class) as a seed-set ($X$) to train the LLP font classifier; the remaining 2,669 documents became the unlabeled set ($U$).

After each step of training, the active learner selected a batch with the 20 most informative documents in $U$ (based on the scoring function), and queried their labels from an oracle[2] that played the role of the human labeler. These newly-labeled documents were added to the training set $L$, the font classifier was retrained, and its performance was tested on the 600 instances in the test set. We repeated this process until all unlabeled data was consumed, recording the size of the labeled data set and classifier accuracy. For comparison, we included a baseline system that iteratively selected a random set of 20 documents. For evaluation purposes, the LLP bandwidth parameter was set to 300. Each experiment was repeated 20 times with a new set of random seeds to get a stable performance estimate.

To ascertain the relative merit and degree of complementarity of the three selection criteria, we report performance for each of them in isolation[3] and for each of their linear combinations:

$$\begin{aligned} S_1(k) &= H(y|U_k, L) \\ S_2(k) &= D_k \\ S_3(k) &= H(y|U_k, L) + D_k \\ S_4(k) &= D_k + D'_k \\ S_5(k) &= H(y|U_k, L) + D'_k \\ S_6(k) &= H(y|U_k, L) + D'_k + D_k \end{aligned}$$

where $H(y|U_k, L)$ is the uncertainty measure, $D'_k$ is the diversity factor, and $D_k$ is the dissimilarity measure.

Learning curves for these scoring functions and the baseline random selection are shown in Figure 6a. The best performer is $S_5$ (uncertainty+diversity), which requires 523 labeled samples (17%) to achieve a maximum average test accuracy of 89%. The uncertainty criterion alone ($S_1$) also performs well, achieving a maximum test accuracy of 88% using 503 labeled samples.

In a final analysis, we calculated the area under the curve (AUC) of each learning curve as a scalar performance measure for each active sampling approach. Results are summarized in Figure 6b. Based on the AUC values, the uncertainty criterion performs marginally better than the sampling based on uncertainty+diversity. The remaining sampling techniques, all of which use dissimilarity, perform notably worse. This may be attributed to the characteristics of our dataset. As shown in Figure 5b, the distribution of BoFs reveals a good degree of separability between Roman and Blackletter fonts, with some overlap with the Mixed class. This arrangement of data makes exploration less useful because most of the information is captured by instances at the class boundaries. As a result, active sampling strategies that involve exploration need more labeled data to reach a desired accuracy compared with a more exploitative technique that samples at the class boundaries.

---

[2] The oracle in this case was the training dataset itself, which had been fully labeled in advance.
[3] We do not consider Diversity in isolation because its primary use is to aid other sampling techniques to pick diverse unlabeled instances.

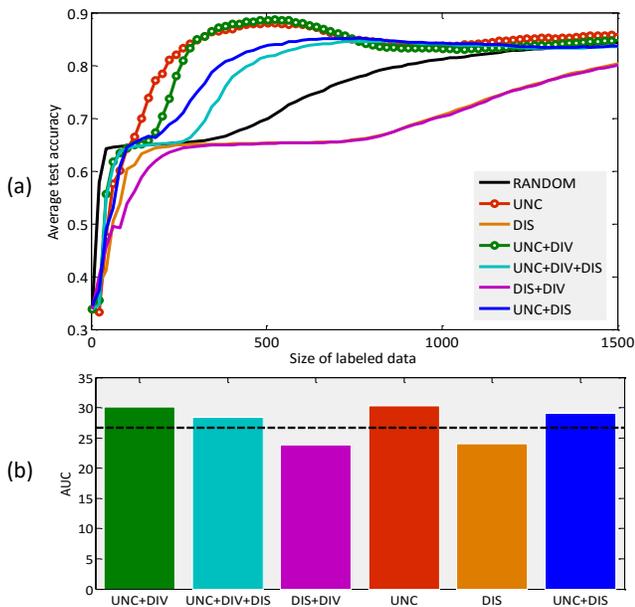

*Figure 6: (a) Learning curves and (b) normalized AUC for different sampling techniques. Black line in (b) denotes random sampling.*

## Discussion

We have developed a font-identification system for historical documents that uses active learning to reduce the amount of labeled data required to train a classifier. The system extracts image-based features that capture differences between Blackletter and Roman font families at the letter and word levels. When tested on a dataset containing 500 Blackletter documents and 500 Roman documents, the system achieves an 84% classification rate at the word level. These features are then combined into a single feature vector for each document page through a bag-of-word features (BoFs) representation, and then used to train logistic label propagation (LLP) model.

We evaluated six different active sampling strategies based on uncertainty, dissimilarity and diversity criteria, and compared them against random active sampling. When tested on a document collection with over 3,000 documents (Blackletter, Roman and Mixed), our results show that a combined uncertainty+diversity criterion can achieve an 89% classification accuracy on test using only a small fraction (17%) of all the training instances. Further analysis (results not shown) reveals that most misclassification errors (79%) occur for Mixed-font documents, with additional errors arising due to italicized Roman fonts with slant line density similar to Blackletter, as well as thin Blackletter fonts with mean and IQR character width comparable to that of Roman fonts.

Our work has focused on font identification but can be easily adapted to identify other problems in historical document collections, such as bleed through, graphic content, musical scripts, and decorative page elements. Further work will also explore the development of graphical user interfaces and interactive machine learning strategies (e.g. Cueflik (Fogarty et al. 2008)) to allow scholars to tag, explore and understand large document collections more efficiently.